%% file: _main.tex
\documentclass[letterpaper, 10 pt, conference]{ieeeconf}  

\IEEEoverridecommandlockouts                              

\input{preamble}

\newcommand{\methodname}{\textsc{Actron3D}\xspace}
\newcommand{\methodnametitle}{\textit{Actron3D}}
\newcommand{\fieldname}{\text{Neural Affordance Function}}

\begin{document}
\title{\LARGE \bf
\textcolor{Periwinkle}{\methodnametitle}: Learning Actionable Neural Functions from Videos for Transferable Robotic Manipulation
}

\renewcommand{\baselinestretch}{0.992}


\author{
  \authorhref{https://dipan-zhang.github.io/}{Anran Zhang}$^{*, 1, 2}$  \quad    
  \authorhref{https://hanzhic.github.io/}{Hanzhi Chen}$^{*, \dagger, 1, 2}$     \quad  
  \authorhref{https://yannickburkhardt.github.io/}{ Yannick Burkhardt}$^{1, 2}$   \quad  
  \authorhref{https://www.linkedin.com/in/yao-zhong-72051223a/}{Yao Zhong}$^{2}$     \\  
  \authorhref{https://scholar.google.com/citations?user=6cuqviYAAAAJ}{Johannes Betz}$^{2}$  \quad
  \authorhref{https://helenol.github.io/}{Helen Oleynikova}$^{1}$  \quad
  \authorhref{https://scholar.google.ch/citations?user=SmGQ48gAAAAJ}{Stefan Leutenegger}$^{1}$ \quad \\
  $^*{\text{ Equal Contribution }}$ $^\dagger{\text{ Project Lead }}$ $^1\text{ ETH Zurich }$ $^2\text{ Technical University of Munich }$
}
\let\oldtwocolumn\twocolumn
\renewcommand\twocolumn[1][]{%
    \oldtwocolumn[{#1}{
    \centering
    \vspace{-12pt}
    \includegraphics[width=1.0\textwidth]{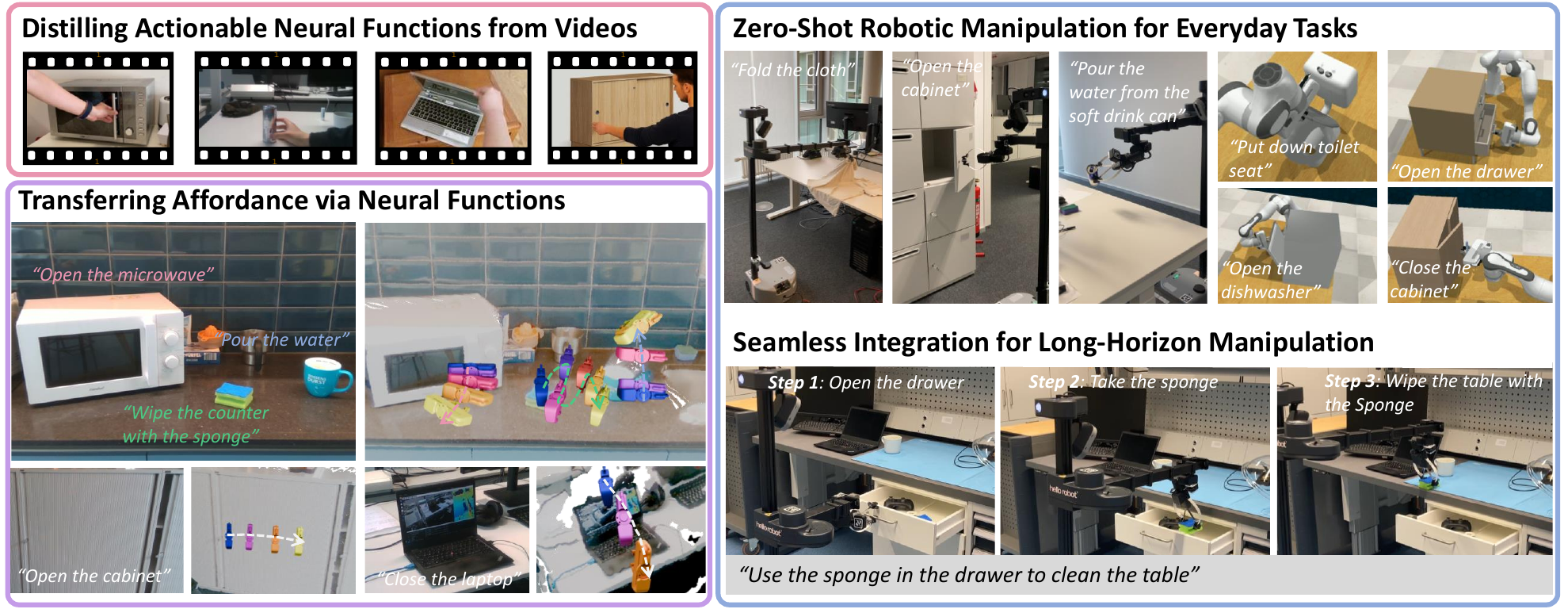}   
    \vspace{-15pt}
    \captionsetup{type=figure}\caption{\small{\textbf{\textcolor{Periwinkle}{\methodnametitle}} is a framework distilling actionable cues from diverse monocular video sources—casual recordings, HOI datasets, or generated contents—into continuous neural representations, enabling zero-shot transfer manipulation skills to novel objects and scenes.
    Project website: \href{https://dipan-zhang.github.io/Actron3D-project/}{\text{https://dipan-zhang.github.io/Actron3D-project/}} 
    }}
    \label{fig:teaser_fig} 
    }]
}
\maketitle

\thispagestyle{empty}
\pagestyle{empty}
\input{sections/00_abstract}
\input{sections/01_intro}
\input{sections/02_literature}
\input{sections/03_method}

\input{sections/04_exp}

\input{sections/06_conclusion}
\input{sections/07_ack}

{
    \footnotesize
    \bibliographystyle{IEEEtran}
    \typeout{}
    \bibliography{reference}
}

\end{document}

%% file: preamble.tex
\usepackage{colortbl}
\usepackage{bm}
\usepackage{amsmath}
\usepackage[T1]{fontenc}
\usepackage{soul}
\usepackage{multirow}
\usepackage{makecell}
\usepackage{booktabs}
\usepackage{nicefrac}
\usepackage{animate}
\usepackage{wrapfig}
\usepackage{algorithm}
\usepackage[noend]{algpseudocode}
\usepackage{amssymb}
\usepackage{etoolbox}
\let\labelindent\relax
\usepackage{enumitem}
\usepackage{graphicx}
\usepackage[font={small}]{caption}
\usepackage{hyperref}
\hypersetup{
    colorlinks=true,
    citecolor=LimeGreen,
    linkcolor=WildStrawberry,
    urlcolor=cyan,
}
\usepackage{listings}
\usepackage{subcaption}
\usepackage{tabularx}
\usepackage{tikz}
\usetikzlibrary{fadings}
\usepackage{titlesec}
\usepackage{xcolor}
\usepackage{soul}
\usepackage{multirow}  
\usepackage[dvipsnames]{xcolor} 

\input{math_commands}

\definecolor{deepblue}{rgb}{0,0,0.5}
\definecolor{deepred}{rgb}{0.6,0,0}
\definecolor{magenta}{rgb}{1.0,0,1.0}
\definecolor{deepgreen}{rgb}{0,0.5,0}
\definecolor{textblue}{rgb}{.2,.2,.7}
\definecolor{textred}{rgb}{0.54,0,0}
\definecolor{textgreen}{rgb}{0,0.43,0}
\definecolor{es-blue}{rgb}{0.1372,0.666,1}
\definecolor{stefan}{rgb}{0.,0.33.,0.0}
\definecolor{hanzhi}{rgb}{0.08,0.33,0.6}
\definecolor{anran}{rgb}{1.0,0.5,0.0}
\definecolor{allcolor}{rgb}{1.0,0.53,0.0}
\definecolor{author}{rgb}{0.2,0.1,0.6}
\definecolor{seoul}{rgb}{0.0,0.71,0.57}

\newcolumntype{Y}{>{\centering\arraybackslash}X}

\usepackage[capitalise, nameinlink]{cleveref}
\makeatletter
\let\NAT@parse\undefined
\makeatother
\usepackage[numbers,sort&compress]{natbib}
\usepackage{xspace}

\newcommand{\authorhref}[3][author]{\href{#2}{\color{#1}{#3}}}

%% file: math_commands.tex
\usepackage{amsfonts,bm}

\def\eqref#1{equation~\ref{#1}}

\def\1{\bm{1}}

\DeclareMathAlphabet{\mathsfit}{\encodingdefault}{\sfdefault}{m}{sl}
\SetMathAlphabet{\mathsfit}{bold}{\encodingdefault}{\sfdefault}{bx}{n}

\DeclareMathOperator*{\argmin}{arg\,min}

%% file: sections/00_abstract.tex
\begin{abstract}
We present \methodname, a framework that enables robots to acquire transferable 6-DoF manipulation skills from monocular, uncalibrated, RGB-only human demonstration videos.
Our key idea is to represent manipulation knowledge within a video as a continuous neural function over object space. At the core of \methodname lies the \textit{\fieldname}, which distills geometry, visual features, contact priors, and action flows from diverse demonstration videos into a compact 3D neural representation. 
During deployment, we adopt a hierarchical pipeline that retrieves the matched affordance function and transfers encoded manipulation knowledge to novel objects through coarse-to-fine differentiable optimization. Leveraging the continuous nature of \fieldname, the framework performs spatial queries over multimodal features to align demonstrations with observations and generates precise 6-DoF manipulation policy. 
Experiments in both simulation and the real-world demonstrate that \methodname significantly outperforms prior methods, achieving a 14.9 percentage point improvement in the average success rate across 13 tasks while requiring only 2–3 demonstration videos per task.
\end{abstract}

%% file: sections/01_intro.tex
\section{Introduction }\label{sec:intro}

Building a generalist robot remains challenging. Such an agent must acquire interaction skills and transfer them to unseen objects and environments across embodiments. In stark contrast to contemporary data-hungry robotics systems, humans provide a powerful example of efficient learning: Infants, for example, can learn tasks such as opening a cabinet by observing only a few demonstrations and readily transfer this knowledge to similar instances~\cite{jones2007imitation}.

Recent approaches~\cite{bahl2023VRB, ju2024roboabc, kuang2024ram} follow this direction by leveraging scalable human videos to predict or transfer affordance trajectories in pixel space. Although they improve sample efficiency and reduce the need for expensive robot demonstrations, the absence of explicit 3D reasoning leads to ambiguity when lifting pixel predictions into physically meaningful actions. 
To overcome this, another line of work extracts explicit 3D actionable cues from human videos, such as point flows~\cite{xu2024flow, bharadhwaj2024track2act} or 3D waypoints~\cite{chen2025vidbot, papagiannis2024r+}. However, these representations are often sparse, task-specific, or require manually specified contact points or goal images. Moreover, large viewpoint discrepancies between human demonstrations and robot observations introduce significant domain shifts, which lead to substantial performance drops. These limitations highlight a key challenge: how to represent actionable knowledge from videos in a form that is spatially grounded, transferable across viewpoints and objects, and directly usable for robotic manipulation.

In this work, we address this challenge by representing manipulation knowledge as a continuous neural function defined over object space. Based on this idea, we propose \methodname, a \emph{distill-then-transfer} framework that distills monocular demonstration video into compact neural representation called \textit{\fieldname} (NAF) and transfers them to novel objects and scenes, achieving zero-shot robotic manipulation with high sample efficiency.
In the \textbf{\emph{distillation}} phase, \methodname learns \textit{\fieldname} that jointly encodes multiple modalities, including geometry, visual appearance, dense visual descriptors, contact priors, and object-centric action flows. This representation enables continuous spatial queries over geometry and affordance cues at arbitrary 3D locations, allowing manipulation trajectories to be transferred through differentiable alignment between demonstrations and observations.
In the \textbf{\emph{transfer}} phase, \methodname retrieves the most relevant neural representation from a memory bank and aligns it with the target object through a coarse-to-fine differentiable affordance optimization process. In contrast to prior work that transfers manipulation skills through one-step feature matching in image space~\cite{ju2024roboabc, kuang2024ram, chen2024funcgrasp}, our method performs iterative alignment between multimodal features in 3D space. This enables the system to naturally handle scale and viewpoint discrepancies and generate spatially grounded 6-DoF manipulation trajectories.

We evaluate \methodname~through extensive simulation and real-world experiments to assess its ability to transfer manipulation skills across objects and viewpoints.
With only 2--3 demonstration videos per task, it outperforms several data-hungry baselines~\cite{chen2025vidbot, yuan2024generalflow, bahl2022human, kuang2024ram, mo2021where2act} by a 14.9 percentage point (pp) improvement in the average success rate across 13 tasks.

In summary, our main contributions are: 
(1) A continuous object-centric action representation (\textit{Neural Affordance} \textit{Function}) that jointly encodes geometry, visual features, contact priors, and object-centric point flows distilled from monocular video.
(2) A differentiable affordance transfer mechanism that aligns neural affordance function with novel scenes to generate executable 6-DoF manipulation trajectories.
(3) Comprehensive experiments and downstream applications in both simulation and real-robot settings that validate the effectiveness and versatility of \methodname.

%% file: sections/02_literature.tex
\section{Related Work} \label{sec:literature}

\noindent \textbf{Robotic Affordance Grounding.}
Affordance refers to the actionable properties of objects, indicating where and how an agent should interact with them. Early works~\cite{Austin2015AffordanceDet, chuang2018learning, do2018affordancenet} focused on learning 2D affordance from human-annotated datasets via end-to-end training. However, their dependence on manual annotation limits scalability and hinders generalization to unseen environments.
To overcome this, recent approaches~\cite{ju2024roboabc, kuang2024ram} leverage out-of-domain human videos to transfer affordance knowledge. However, these methods operate solely in the 2D image plane and lack 3D spatial grounding. Moreover, they typically rely on large, manually curated memory banks, increasing system complexity.
Another direction~\cite{mo2021where2act, wu2022vatmart, wang2022adaafford} explores 3D affordance learning in simulation. Although more spatially expressive, these methods require extensive synthetic assets and suffer from sim-to-real transfer challenges.
In contrast, our method infers spatially grounded affordance using a compact memory bank of casually captured or generated videos, enabling zero-shot transfer without extensive annotations or simulation.

\noindent \textbf{Visual Features for Robotic Manipulation.}
Recent advances have employed visual descriptors from foundation models to enable language-guided and context-aware robotic manipulation. A prominent line of work uses these descriptors for keypoint-based object correspondence~\cite{ju2024roboabc, kuang2024ram, florence2018dense}, supporting generalization from in-the-wild human demonstrations. However, the lack of 3D spatial grounding makes these methods sensitive to viewpoint changes.
To improve robustness, several approaches~\cite{rashid2023lerftogo, shen2023F3RM, wang2024sparsedff, ze2024gnfactor, wang2024d3fields} lift 2D features into 3D fields via differentiable rendering across multi-view captures. These 3D-grounded features serve as spatially consistent intermediates for policy learning or as matching costs for action planning. Yet, the reliance on controlled multi-view setups limits their scalability to unconstrained environments.
In contrast, leveraging recent advances in novel view synthesis~\cite{wonder3d_long2023, xu2024instantmesh}, our approach reconstructs coherent 3D fields from monocular videos. These fields jointly encode multimodal action-related features, enabling robust affordance transfer to novel objects and viewpoints without requiring complex capture setups.

\noindent \textbf{Robot Learning from Videos.} Previous work has explored various video sources to guide robot learning of manipulation skills. One line of research uses human videos to learn visual representations~\cite{nair2022r3m, xiao2022masked, xu2024flow} or reward functions~\cite{bahl2022human, smith2019avid}, thus facilitating the learning of visuomotor policies. Another line utilizes MoCap systems to re-target human motions into the robot's action space~\cite{qin2022dexmv, wang2024dexcap, shaw2023videodex, papagiannis2024r+}. However, these approaches are often limited to controlled lab settings due to infrastructure requirements. More recent methods attempt to infer affordance directly from large-scale human videos on the web~\cite{bahl2023affordances, yuan2024generalflow, bharadhwaj2024track2act, chen2025vidbot}, but they face several limitations: they often require manual specification of contact points or goal images, or reduce interactions to sequences of 3D waypoints. Such simplifications make it difficult to perform more dexterous tasks, such as pouring water. Recent advances in generative video models have enabled robots to visually imagine actions, thereby expanding the types of video data that can be used for policy learning. For example, works like~\cite{liang2024dreamitate, bharadhwaj2024gen2act, patel2025robotic} infer robot policies conditioned on synthesized videos of human interactions. However, video generation at test time can be highly computationally expensive. In contrast, we propose a scalable framework to distill manipulation skills from diverse videos into several object-centric representations, enabling the robot to efficiently retrieve 6-DoF interaction trajectories from these representations without extensive demonstration or generation.

%% file: sections/03_method.tex
\section{Method}\label{sec:method}
\begin{figure*}[t]
\centering
\includegraphics[width=\linewidth]{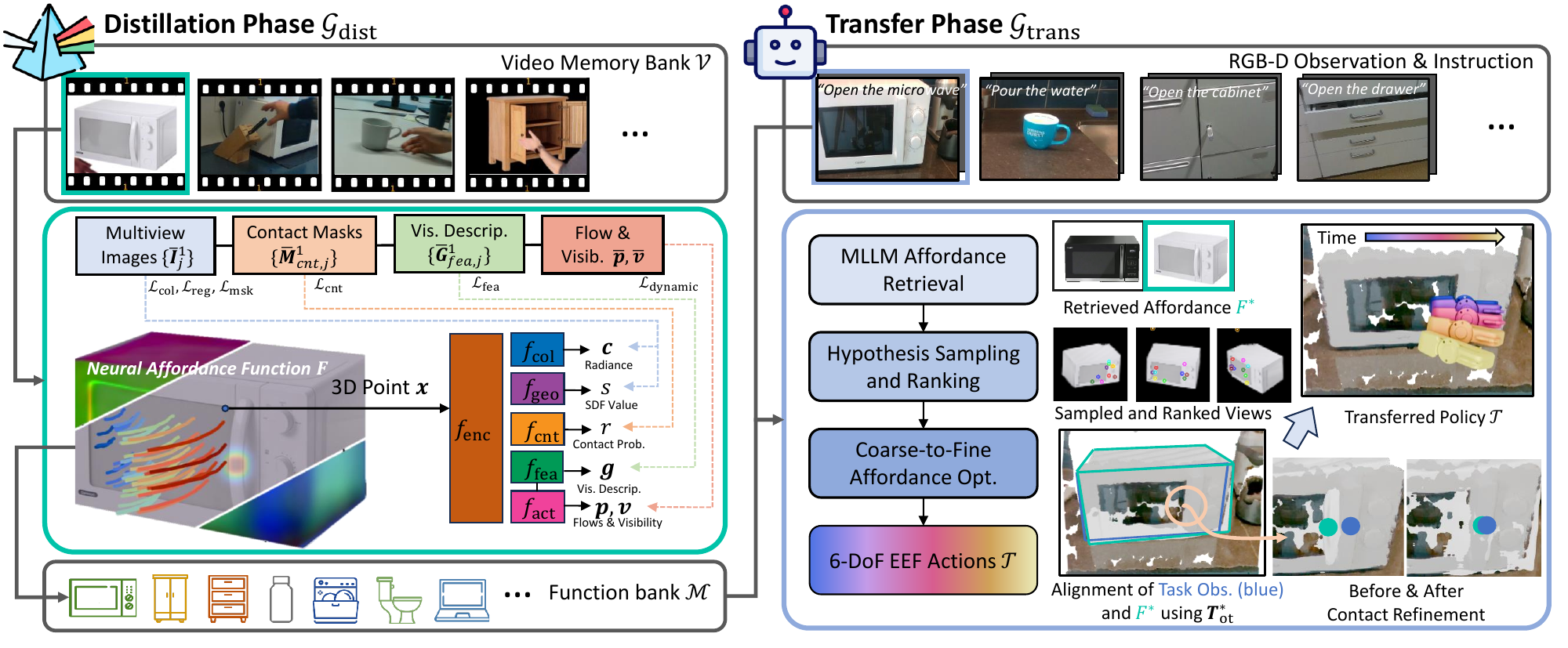}
\vspace{-0.2in}
\caption{Overview of \textbf{\methodname} framework. In the distillation phase, multimodal knowledge from RGB videos is encoded into \fieldname\text{s}, and forming a sparse function bank $\mathcal{M}$. At inference time, $\mathcal{G}_\mathrm{trans}$ retrieves the best matched NAF, aligns it to the target object using a coarse-to-fine optimization module and produces 6-DoF EEF actions (Sect.~\ref{sec:transfer}), in a zero-shot manner.}
\label{fig:pipeline}
\vspace{-0.2in}
\end{figure*}

\subsection{Problem Formulation} 
To learn robot manipulation policies from monocular videos and transfer them to unseen environments, our method consists of two modules: (1) an affordance function distillation module \(\mathcal{G}_\mathrm{dist}\) and (2) an affordance transfer module $\mathcal{G}_\mathrm{trans}$. In the distillation stage, 
\(\mathcal{G}_\mathrm{dist}\) converts each language-narrated demonstration video $(\hat{V}_i, \hat{l}_i)$ into \textit{\fieldname} (\textsc{NAF}) \(F_i = \mathcal{G}_\mathrm{dist}(\hat{V}_i, \hat{l}_i)\), which encodes geometry, visual features, contact priors and actionable motion (Sect.~\ref{sec:naf}). These functions are stored in a function bank \(\mathcal{M} = \{F_i\}_{i=1}^{N_\text{V}}\). At inference time, given a task described by an RGB-D observation and language instruction \((\bm{I}^\text{t}, \bm{D}^\text{t}, l^\text{t})\), \(\mathcal{G}_\mathrm{trans}\) retrieves the most relevant affordance function from $\mathcal{M}$, aligns it with the target scene with affordance optimization, and yields a manipulation policy: $\mathcal{T} = \mathcal{G}_\mathrm{trans}(\bm{I}^\text{t}, \bm{D}^\text{t}, l^\text{t}, \mathcal{M})$ (Sect.~\ref{sec:transfer}), where the output policy $\mathcal{T}$ is a sequence of 6-DoF EEF actions over a planning horizon \(H\).


\subsection{Preliminaries}
\noindent \textbf{Neural Signed Distance Function (NeuS).}
We build our \textsc{NAF} based on NeuS~\cite{wang2021neus} and adopt an SDF-based geometric representation instead of 3D Gaussian Splats~\cite{kerbl3Dgaussians} for more accurate geometry reconstruction~\cite{Drawer2025xia}, which serves as the geometric backbone of \textsc{NAF}.
Formally, a \textit{signed distance function} (SDF) \( \Omega: \mathbb{R}^3 \rightarrow \mathbb{R} \) maps a 3D point \(\bm{x}\) to its signed distance from the object surface:
\(
S = \left\{ \bm{x} \in \mathbb{R}^3 \mid \Omega(\bm{x}) = 0 \right\},
\)
enabling realistic object modeling.
NeuS uses standard volume rendering to render any attribute along a ray $\mathbf{r}(s)$, using the weight \(w(s)\) derived from the opaque density \(\rho(s)\) and the accumulated transmittance \(T(s)\) between the near plane \(s_n\) and \(s\).
\begin{align}
    w(s) = T(s) \, \rho(s),
    ~T(s) = \exp\left( - \int_{s_n}^s \rho(u) \, \mathrm{d}u \right).
\label{eq:volume_rendering_weight}
\end{align}

\noindent \textbf{Point Flows as Action Representation.} Instead of directly modeling robot end-effector (EEF) actions, we adopt object-centric 3D point flows as action representation, making this representation embodiment-agnostic, informative and transferable \cite{bharadhwaj2024track2act, xu2024flow}. The point flows encode the rigid displacement of the manipulated object over time, from which the underlying rigid EEF actions can be estimated. We denote point flows in the object frame as \(\bar{\bm{P}} \in \mathbb{R}^{N_\text{f} \times H \times 3} \), where \(N_\text{f}\) stands for the number of points, \(H\) represents the horizon of flow.

\noindent \label{method:6D_action}\textbf{6-DoF Manipulation Policy from Flows.} 
\noindent \textbf{6-DoF Manipulation Policy from Flows.}
Given object-centric point flows $\bm{\bar{P}}\in\mathbb{R}^{N_\text{f}\times H\times3}$, we derive the corresponding EEF trajectory by estimating the motion of the rigid object over time.
Let $\bm{p}_{n,t}\in\mathbb{R}^3$ denote the displacement of the surface point $n$ of timestep $t$.
For each timestep $t$, we form the set of object keypoints
$\bm{p}_t=\{\bm{p}_{n,t}\}_{n=1}^{N_f}\in\mathbb{R}^{N_f\times3}$.

The transformation between consecutive timesteps is obtained by solving
\(
\bm{T}^* =
\arg\min_{\bm{T}\in \mathrm{SE}(3)}
\sum_{n=1}^{N_f}
w_n
\|\bm{p}_{n,t+1}-\bm{T}\bm{p}_{n,t}\|^2,
\)
where $w_n=\frac{1}{d_n+\beta}$ weights points based on their distance $d_n$ from the initial EEF position. The weighted SVD algorithm~\cite{BeslSVD} yields a sequence of relative transformations $\mathcal{T}_{\text{rel}}=\{\bm{T}_t^*\}_{t=1}^{H-1}$. The absolute EEF trajectory $\mathcal{T}$ is obtained by composing $\mathcal{T}_{\text{rel}}$ starting from the initial EEF pose.

\subsection{Neural Affordance Functions Distillation from Videos}
\label{sec:naf}
\noindent \textbf{Formulation.} We propose a 3D object-centric action representation called \textit{\fieldname}, which jointly encodes geometry, color, dense visual features, and affordance cues within a unified continuous function. Formally, $F$ is defined in a canonical object frame and parameterized by the following components with multiple MLPs:
\begin{equation}
\begin{aligned}
F &: \mathbb{R}^3 \rightarrow 
{\color{Purple} \mathbb{R}} \times 
{\color{NavyBlue}\mathbb{R}^3} \times 
{\color{ForestGreen}\mathbb{R}^d} \times {\color{BurntOrange}\mathbb{R}} \times ({\color{RubineRed}\mathbb{R}^{H\times{3}}} \times 
{\color{RubineRed}\mathbb{R}^{H}}),  \\[1mm]
F&(\bm{x}) = \big( f({\color{Brown}f_\texttt{enc}}(\bm{x}), \bm{x}) \;\big|\; f \in \mathcal{F} \big), \quad \bm{x}\in\mathbb{R}^3, \\[1mm]
\mathcal{F} &= \big\{
{\color{Purple}f_\texttt{geo}}, {\color{NavyBlue}f_\texttt{col}}, 
{\color{ForestGreen}f_\texttt{fea}}, {\color{BurntOrange}f_\texttt{cnt}},  
{\color{RubineRed}f_\texttt{act}}
\big\}.
\label{eq:repr_formulation}
\end{aligned}
\end{equation}
For a 3D point $\bm{x}$, the encoder head ${\color{Brown}f_\texttt{enc}}$ produces a latent feature $\bm{z}\in\mathbb{R}^{d_z}$, which serves as the shared geometry backbone for other heads. Conditioned on $\bm{z}$, the geometry head ${\color{Purple}f_\texttt{geo}}$ predicts the signed distance $s\in\mathbb{R}$, and the color head ${\color{NavyBlue}f_\texttt{col}}$ outputs the RGB radiance $\bm{c}\in\mathbb{R}^3$. Although less critical for manipulation, color helps enforce geometry supervision during reconstruction. The feature head ${\color{ForestGreen}f_\texttt{fea}}$ generates $d$-dimensional visual descriptors $\bm{g}\in\mathbb{R}^d$~\cite{oquab2023dinov2} for robust affordance knowledge retrieval and alignment, while the contact head ${\color{BurntOrange}f_\texttt{cnt}}$ estimates the probability $r\in\mathbb{R}$ that $\bm{x}$ lies in a feasible contact region. Notably, both ${\color{ForestGreen}f_\texttt{fea}}$ and ${\color{BurntOrange}f_\texttt{cnt}}$ take $\bm{x}$ in addition to $\bm{z}$ as input. The action head ${\color{RubineRed}f_\texttt{act}}$, defined on the object surface $S$, maps each surface point $\bm{x}^\prime \in S$, augmented with its queried geometric feature $\bm{z}^\prime={\color{Brown}f_\texttt{enc}}(\bm{x}^\prime)$ and visual feature $\bm{g}^\prime={\color{ForestGreen}f_\texttt{fea}}(\bm{x}^\prime)$, to $H$-step flows $\bm{p}\in\mathbb{R}^{H\times3}$ with visibility scores $\bm{v}\in\mathbb{R}^H$.
Hence, we rewrite the $\color{RubineRed}f_\texttt{act}$ defined in Eq.~\ref{eq:repr_formulation} as: $(\bm{p}(x),\bm{v}(x))={\color{RubineRed}f_\texttt{act}} (\bm{z}^\prime,\bm{g}^\prime, \bm{x}^\prime)$. 
Restricting ${\color{RubineRed}f_\texttt{act}}$ to the surface avoids volumetric redundancy and ensures a compact representation of the motion knowledge. 
Conditioning action flows on both geometric and visual features increases distinctiveness and mitigates collapse under imbalanced supervision, since valid flows occur on a small subset of the space.
Unlike previous 2D affordance representations without spatial awareness~\cite{ju2024roboabc,kuang2024ram,bahl2023VRB}, \textsc{NAF} operates directly in 3D, eliminating image-space ambiguity and enabling differentiable affordance optimization via continuous queries of $F$.

\noindent \textbf{Multimodal Differentiable Rendering.} \textsc{NAF} enables rendering any modality defined in volumetric space at arbitrary viewpoint using volume rendering: 
\begin{equation}
\bm{Q}_{{m}, \bm{u}} = \int_{s_\mathrm{n}}^{s_\mathrm{f}} w(s) f_{m}(\bm{r}(s))  \mathrm{d}s.
\label{eq:volume_rendering}
\end{equation} 
where $\bm{r}(s)=\mathbf{o}+s\mathbf{d}$ and \( w(s) \) is the volumetric weight at depth $s$ along the ray. $\bm{Q}_{{m}, \bm{u}}$ is the 2D rendering of modality $m$, where $m$ indexes the rendered modality, i.e., $m \in \big\{ {
{\color{Purple}\texttt{geo}}}, {\color{NavyBlue}\texttt{col}}, 
{\color{ForestGreen}\texttt{fea}}, 
{\color{BurntOrange}\texttt{cnt}}
\big\}$. 
For a pixel $\bm{u}$, we cast a ray $\bm{r}(s)$ sampling modality field $f_{m}(\bm{r}(s))$ from near to far planes $s_\mathrm{n}$ and $s_\mathrm{f}$. 
Such a unified rendering scheme makes it possible to compare different modalities consistently across viewpoints, a key to bridging instance and viewpoint gaps.

\noindent \textbf{Multimodal Knowledge Extraction.} 
Given a narrated demonstration video \( \hat{V}\), we first capture the geometry and appearance of the object. The first hands-free frame \(\bar{\bm{I}}^{1}\) is processed with an image-to-3D model~\cite{wonder3d_long2023} to generate six multi-view images \(\{\bar{\bm{I}}^1_j\}_{j=1}^6\) and corresponding foreground masks \(\{\bar{\bm{M}}^1_j\}_{j=1}^6\)~\cite{kirillov2023segment}. Dense features \(\{\bar{\bm{G}}^{1}_{\texttt{fea}, j}\}_{j=1}^6\) extracted using a pretrained DINO model~\cite{oquab2023dinov2}.

To extract affordance cues, i.e., contact regions and point flows, we process the video \(\hat{V}\) through a SfM pipeline~\cite{li2024_MegaSaM} and a metric depth estimator~\cite{piccinelli2025unidepthv2}, producing per-frame camera extrinsics and dense depth maps. Following prior work~\cite{bahl2023VRB, ju2024roboabc}, we identify the first contact frame \(k_{\text{c}} > 1\) and its contact regions. The contact regions are warped to the first frame using recovered camera parameters and depth maps, yielding contact masks aligned with the first view and its synthesized novel views \(\{\bar{\bm{M}}^1_{\texttt{cnt}, j}\}_{j=1}^6\).  
We track \(N_\text{f}\) keypoints sampled within the object's foreground mask using a 3D point tracker~\cite{tapip3d}. This yields 3D point flows expressed in the coordinate frame of the first camera view, along with the corresponding visibility:
\(
{_\text{c}}\bm{P} \in \mathbb{R}^{N_\text{f} \times H \times 3},
\bm{V} \in \mathbb{R}^{N_\text{f} \times H}.
\)

To transform the extracted flows \( {}^{\text{c}}\bm{P} \) into the frame of \textsc{NAF}, we estimate a similarity transformation \(\bm{T}_{\text{oc}} \in \mathrm{SIM}(3)\) via optimization. The optimal pose is obtained by minimizing SDF values over the transformed object points:
\(
\bm{T}_\text{oc}^{*} = \argmin_{\bm{T}_\text{oc}} \sum_{\bm{x} \in  {_\text{c}}\mathcal{X}}   \big\| {\color{Purple}{f_\texttt{geo}}}( {\color{Brown}f_\texttt{enc}}( \bm{T}_\text{oc} \bm{x}), \bm{T}_\text{oc} \bm{x}) \big\|.
\label{eq:sdf_pose_selection}
\)
Finally, the extracted flow in the object's space is denoted as: 
\(
\bar{\bm{P}} = \bm{T}_{\text{oc}}^* {_\text{c}}\bm{P}.
\)
To obtain supervision over the surface of the object, we further sample points $\bm{x}\in S$ and assign flows from nearby tracked points using nearest-neighbor interpolation, producing dense surveillance signals $\bar{\bm{p}}(\bm{x})$ and $\bar{\bm{v}}(\bm{x})$.

\noindent \textbf{Neural Function Fitting.}
The fitting process is split into two phases: first, we obtain geometric and visual understanding (\emph{static phase}); then, we extract the object's surface to constrain action flows within a bounded space (\emph{dynamic phase}).

\textbf{\emph{Static Phase.}} All heads except the action are fitted with:
\begin{equation}
\label{eq:geometry_loss}
\mathcal{L}_\mathrm{static} = \mathcal{L}_\text{neus} + \lambda_{3} \mathcal{L}_{\mathrm{fea}} + \lambda_{4} \mathcal{L}_\mathrm{cnt},
\end{equation}
where \(\mathcal{L}_\text{neus}=\mathcal{L}_{\mathrm{col}} + \lambda_{1} \mathcal{L}_{\mathrm{reg}} + \lambda_{2} \mathcal{L}_{\mathrm{msk}} \) follows NeuS~\cite{wang2021neus}. We render $\bm{Q}_{\texttt{fea},\bm{u}}, \bm{Q}_{\texttt{cnt},\bm{u}}$ via Eq.~\ref{eq:volume_rendering} and compare them with the labels $\bar{\bm{G}}_{\texttt{fea},\bm{u}}, \bar{\bm{M}}_{\texttt{cnt},\bm{u}}$ for each pixel $\bm{u}$ sampled from the image plane $\mathcal{U}$. $\mathcal{L}_{\mathrm{cnt}}$ is computed with the binary cross-entropy (BCE) loss.
\begin{align}
\mathcal{L}_{\mathrm{fea}} &= \sum_{\bm{u} \in \mathcal{U}} 
\big\|\bar{\bm{G}}_{\texttt{fea}, \bm{u}}-  \bm{Q}_{\texttt{fea}, \bm{u}}\|^2, 
\label{eq:featloss}\\
\mathcal{L}_{\mathrm{cnt}} &= \sum_{\bm{u}\in \mathcal{U}} 
\operatorname{BCE}\big(\bar{\bm{M}}_{\texttt{cnt}, \bm{u}},\bm{Q}_{\texttt{cnt}, \bm{u}}\big).
\label{eq:contactloss}
\end{align}

\textbf{\emph{Dynamic Phase.}} With the other heads frozen, the action head is fitted using:
\begin{equation}
\label{eq:flow_loss}
\mathcal{L}_{\mathrm{dynamic}} = \sum_{\bm{x} \in S} \big\| \bm{\bar{p}}(\bm{x}) - \bm{p}(\bm{x}) \|^2 + \big\| \bm{\bar{v}}(\bm{x}) - \bm{v}(\bm{x}) \|^2,
\end{equation}
where \(\bm{\bar{p}}, \bm{\bar{v}}\) are the flows and visibility of ground-truth, and \(\bm{p}, \bm{v}\) are the predictions of the action head $\color{RubineRed}f_\texttt{act}$.

\begin{figure}[t]
\centering
\includegraphics[width=\linewidth]{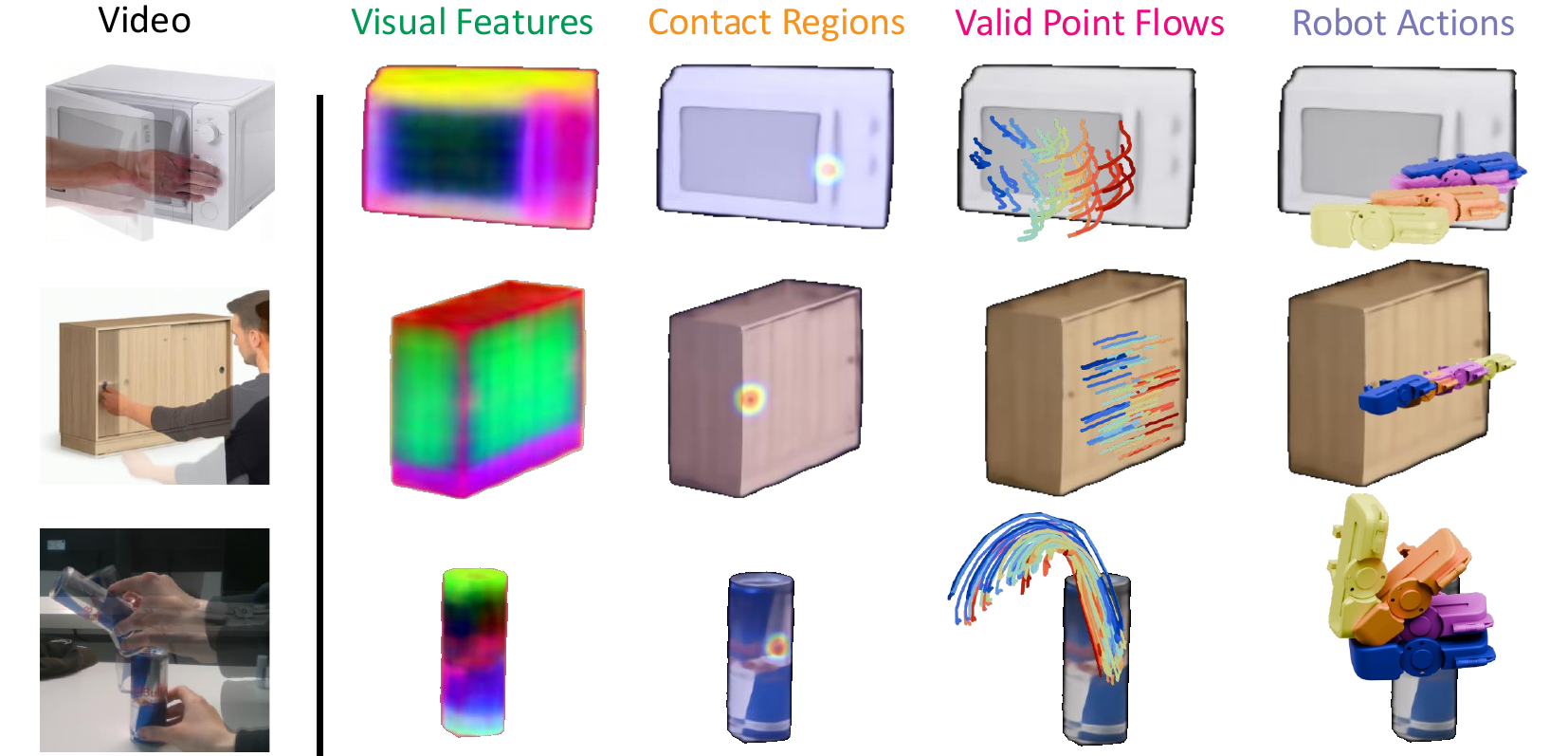}
\vspace{-0.2in}
\caption{Multimodal information and robot actions synthesized from novel views of \textsc{NAF}s fit to diverse videos.}
\label{fig:NAF_Novel_views}
\vspace{-0.3in}
\end{figure}

\subsection{Affordance Transfer via Neural Functions}
\label{sec:transfer}
Each fitted neural affordance function ${F}$ is paired with a language narration $\hat{l}$ and stored in the function bank $\mathcal{M}$. At inference time, given a new task specified by an RGB-D observation and a natural-language instruction $(\bm{I}^\text{t}, \bm{D}^\text{t}, l^\text{t})$, we retrieve the most relevant affordance function ${F}^*$ from $\mathcal{M}$ and transfer its encoded action knowledge to the target scene, as shown in Fig~\ref{fig:pipeline} $\mathcal{G}_\text{trans}$.
Transfer is performed by affordance optimization, estimating a $\mathrm{SIM}(3)$ transformation $\bm{T}_{\text{ot}}^*$ that jointly accounts for scale and pose while aligning ${F}^*$ with the task observation geometrically and semantically by querying ${\color{Purple}f_\texttt{geo}}$, ${\color{ForestGreen}f_\texttt{fea}}$ and ${\color{BurntOrange}f_\texttt{cnt}}$.
Once aligned, ${\color{BurntOrange}f_\texttt{cnt}}$ and ${\color{RubineRed}f_\texttt{act}}$ of $F^*$ can be directly queried and converted into executable actions $\mathcal{T}$. We demonstrate several distilled \textsc{NAF}s in Fig.~\ref{fig:NAF_Novel_views}.

\noindent \textbf{Affordance Knowledge Retrieval.} 
We use a Multimodal Large Language Model (MLLM)~\cite{openai2024gpt4o} to retrieve the best-matched \textsc{NAF} ${F}^*$, comparing task observation and instruction with \textsc{NAFs} in the function bank. Unlike the CLIP-based retrieval used in previous work~\cite{kuang2024ram} which computes visual and textual similarities in separate embedding spaces, MLLM performs joint reasoning over paired multimodal inputs. For each stored \textsc{NAF} ${F}_i$, we construct a descriptor $\mathcal{O}_i = \{\bar{\bm{I}}^1_i, \hat{l}_i\}$ consisting of each video's first hand-free frame and its associated narration. We batch all descriptors $\{\mathcal{O}_i\}_{i=1}^{N_\text{V}}$ as candidate and prompt the MLLM to select the one most consistent with the give task query. This formulation allows for exploiting both fine-grained visual appearance (e.g., shape) and high-level linguistic context (e.g., "open the \textit{top} drawer" versus "open the \textit{bottom} drawer"), enabling more precise retrieval across object instances and affordance types.

\noindent \textbf{Contact-Guided Hypothesis Sampling and Ranking.}
From the selected \textsc{NAF} ${F}^*$, we sample and rank coarse view hypotheses to initiate the optimization process. Around the origin of the canonical object frame associated with ${F}^*$, we sampled \( N_{\text{r}} \) evenly spaced viewpoints at a fixed radius and elevation. Viewpoints with occluded contact regions are discarded. The remaining feasible candidates are ranked by visual feature consistency with the target observation. For each viewpoint, we render its feature map ${\bm{Q}}_\texttt{fea}$ encoded in ${\color{ForestGreen}f_\texttt{fea}}$ using Eq.~\ref{eq:volume_rendering} and establish dense correspondences with the target feature map ${\bm{G}}^\text{t}_\texttt{fea}$, extracted from $\bm{I}^\text{t}$ with same DINO model, using best-buddy matching~\cite{amir2021deep}. We then compute the average squared correspondence distances as the geometric consistency metric. The top-\( k \) hypotheses with the lowest distances are selected for the subsequent pose optimization stage.

\noindent \textbf{Affordance Optimization.} This coarse-to-fine affordance optimization first achieves global alignment before refining task-specific contact regions, ensuring alignment between observation and demonstration. Direct optimization of contacts often leads to local minima; therefore, we initially perform coarse alignment on the top-$k$ candidates with 
\(
\mathcal{L}_\mathrm{coarse} = \mathcal{L}_\mathrm{fea} + 
\beta_\mathrm{1}\mathcal{L}_\mathrm{surf} +
\beta_\mathrm{2}\mathcal{L}_\mathrm{depth} 
\label{eq:coarse_opt} 
\):
\begin{align}
\mathcal{L}_\mathrm{fea} &= \sum_{\bm{u}\in \mathcal{U}} \big( \bm{1} - \cos\left({\bm{G}}_{\texttt{fea}, \bm{u}}^\mathrm{t}, \bm{Q}_{\texttt{fea}, \bm{u}}\right) \big),
\label{eq:feature_loss} \\
\mathcal{L}_\mathrm{surf} &= \sum_{\bm{x} \in {_{\mathrm{t}}\mathcal{X}}} \mathcal{L}_\mathrm{huber}\left({\color{Purple}{f_\texttt{geo}}}\left( {\color{Brown}f_\texttt{enc}}( \bm{T}_\mathrm{ot} \bm{x}), {\bm{T}_{\mathrm{ot}}\bm{x}} 
 \right), \bm{0}\right),
\label{eq:sdf_loss} \\
\mathcal{L}_\mathrm{depth} &= \sum_{\bm{u}\in \mathcal{U}}  \mathcal{L}_\mathrm{huber}\left( {\bm{D}}_{\bm{u}}^\mathrm{t}, \bm{D}_{\bm{u}}\right).
\label{eq:depth_loss} 
\end{align}
The feature term $\mathcal{L}_\mathrm{fea}$ is defined as the cosine similarity loss between the feature map $\bm{Q}_{\texttt{fea}, \bm{u}}$ rendered by ${\color{ForestGreen}f_\texttt{fea}}$ and the target feature map ${\bm{G}}^\mathrm{t}_{\texttt{fea}, \bm{u}}$.
The surface term $\mathcal{L}_\mathrm{surf}$ queries ${\color{Purple}{f_\texttt{geo}}}$ for the SDF values of the transformed point cloud ${\bm{T}_{\mathrm{ot}}}\bm{x}$ obtained from the RGB-D observation. The depth term $\mathcal{L}_\mathrm{depth}$ measures the discrepancy between the rendered depth map $\bm{D}_{\bm{u}}$ and the observed depth ${\bm{D}}_{\bm{u}}^\mathrm{t}$ at the same pixel $\bm{u}$.
The coarse pose $\bm{T}_{\mathrm{ot}}^\prime$ is obtained by minimizing the following energy function:
\(
\bm{T}_{\mathrm{ot}}^\prime = \arg\min_{\bm{T}_{\mathrm{ot}}} \mathcal{L}_\mathrm{coarse}.
\label{eq:sdf_pose_selection2}
\) 

Then, in the fine stage, we inject extra contact refinement while preserving the feature-metric and geometric alignment:
\(
\mathcal{L}_\mathrm{fine} = \mathcal{L}_\mathrm{fea} +  \beta_\mathrm{3}\mathcal{L}_\mathrm{cnt} + \beta_\mathrm{4}\mathcal{L}_\mathrm{surf}
\), where $\mathcal{L}_\mathrm{cnt} $ is defined as: 
\begin{align}
\mathcal{L}_\mathrm{cnt} &= \sum_{\bm{q} \in {\mathcal{Q}}^\mathrm{t}_\text{c}} \min_{\bm{q}^\prime \in \mathcal{Q}_\mathrm{c}}\| \bm{q}^\prime  - \bm{T}^{\prime}_{\mathrm{ot}}\bm{q}  \|.
\end{align}
${\mathcal{Q}}_\text{c}$ denotes the set of contact points extracted using ${\color{BurntOrange}f_\texttt{cnt}}$ above the contact probability threshold $\delta_\text{cnt}$, and $\mathcal{Q}^\mathrm{t}_\mathrm{c} \subset {_{t}\mathcal{X}}$ represents the matches based on the cosine similarity of the visual features. The final alignment initialized with $\bm{T}_{\mathrm{ot}}^\prime$ is given by
\(
\bm{T}_{\text{ot}}^* = \argmin_{\bm{T}^\prime_{\text{ot}}} \mathcal{L}_\text{fine}
\label{eq:final_opt}
\). 
Notably, the formulation is modular and allows additional constraints (e.g., collision avoidance) to be incorporated into the energy.

With the final alignment $\bm{T}_{\text{ot}}^*$ established, we query the action head ${\color{RubineRed}f_\texttt{act}}$ on the transformed target point cloud $\bm{T}_{\text{ot}}\,{_{t}\mathcal{X}}$.
For each surface point $\bm{x}$, the network predicts a flow and visibility $(\bm{p}(\bm{x}),\bm{v}(\bm{x})) = {\color{RubineRed}f_\texttt{act}}(\bm{x})$. Points with positive visibility are selected and their predicted flows $\bm{p}(\bm{x})$ are aggregated to estimate a sequence of relative end-effector poses $\mathcal{T}_\text{rel}$ as described in Sect.~\ref{method:6D_action}.  
For robot execution, we sample the contact points in ${\mathcal{Q}^\mathrm{t}}_\text{c}$ and employ a grasp detector~\cite{fang2023anygrasp} to detect a 6-DoF grasp pose around them. The absolute end-effector poses $\mathcal{T} = \{\bm{T}_i\}_{i=1}^H$ are composed of $\mathcal{T}_\text{rel}$ and the detected grasp pose for execution.


\subsection{Implementation Details}
\noindent\textbf{Network Architecture.} Heads ${\color{Brown}f_\texttt{enc}}$, $\color{Purple}f_\texttt{geo}$, and ${\color{NavyBlue}f_\texttt{col}}$ share identical architectures as~\cite{wang2021neus}, with $d_z$ = 256.
The feature head ${\color{ForestGreen}f_\texttt{fea}}$ has an additional 12-level hash-grid encoding~\cite{mueller2022instant} and a 6-band positional encoding, followed by a 2-layer MLP for decoding. 
The contact head ${\color{BurntOrange}f_\texttt{cnt}}$ uses the same positional encoding as ${\color{ForestGreen}f_\texttt{fea}}$, followed by a 3-layer MLP for decoding.
Finally, the action head ${\color{RubineRed}f_\texttt{act}}$ takes surface points with a 10-band positional encoding, fuses them with $\bm{g}$ and $\bm{z}$ (compressed to dimension 96 by a 2-layer MLP), and processes the concatenated representation with a 3-layer MLP. All MLP modules have hidden dimension 128.

\noindent\textbf{Fitting Protocol.}
In the \textit{static phase}, we jointly optimize geometry, color, features, and contact heads using Adam~\cite{kingma2014adam} with the learning rate $5\times10^{-4}$. The loss weights are $\lambda_1=\lambda_3=0.1$, $\lambda_2=1.0$, $\lambda_4=0.5$. This stage takes 3k iterations and roughly 5 minutes. In the \textit{dynamic phase}, the action head is trained while freezing the static components for an additional 1k steps ($\sim$ 3 minutes) with the learning rate $1\times10^{-3}$. For each manipulation task, we require only 2–3 demonstration videos.

\noindent\textbf{Inference Protocol.} We select the top $k=3$ poses and use an Adam optimizer with learning rate $1\times10^{-2}$ and weights $\beta_1=\beta_4=1000, \beta_2=\beta_3=100$ for affordance optimization. The coarse and refinement stages run for 100 and 50 iterations each ($\sim$ 25 seconds in total).

%% file: sections/04_exp.tex
\section{Experiments} \label{sec:exp}
\input{tables/quantiative_results}

Here, we demonstrate the following aspects of our method:
1) It outperforms several strong baseline models across various household tasks in both simulator and real-world settings. 
2) Representing actionable knowledge as a continuous function provides a more feasible and intuitive robot motion, leading to better performance. 
3) Compared with 2D methods, the representation of 3D neural functions achieves a performance gain by a large margin.
4) It can be seamlessly deployed for several downstream applications. 

\subsection{Experiment Setup}
\noindent \textbf{Simulator Setup.} 
We use RLBench~\cite{james2019rlbench} as a simulation environment, deploying a \emph{Franka-Emika Panda} to interact with target objects. Our evaluation covers 13 RLBench tasks that involve manipulating both articulated and portable objects. For each task, we assess the performance under three different viewpoints, generating five trajectories per viewpoint, resulting in a total of 15 trials per method. Following previous work~\cite{chen2025vidbot, kuang2024ram,yuan2024generalflow}, we report the success rate \% (SR) as the primary evaluation metric. A trial is considered successful if the robot's end-effector manipulates the target object along the intended DoF beyond a predefined threshold.

\noindent \textbf{Real Robot Setup.} We conduct real-world manipulation experiments in previously unseen human-suited environments. The test platform is a \emph{Hello Robot Stretch 3} 
equipped with an on-board RGB-D camera for perception. 

\noindent \textbf{Baseline Models.} 
We compare against five representative baselines for 2D/3D affordance prediction. \textbf{Where2Act}~\cite{mo2021where2act} learns 3D actionable affordance from simulated articulated objects, while \textbf{VRB}~\cite{bahl2023VRB} trains on large-scale human videos to predict 2D affordance on the image plane. \textbf{RAM}~\cite{kuang2024ram}, similar to our method, follows a retrieve-and-transfer scheme. However, it retrieves 2D affordances from a significantly larger demonstration bank and requires lifting pixel-wise 2D affordance to 3D using local geometric priors. \textbf{GFlow}~\cite{yuan2024generalflow} and \textbf{VidBot}~\cite{chen2025vidbot} both learn 3D affordance from egocentric videos; \textbf{GFlow} outputs full 6-DoF poses but requires manual input of contact regions, while \textbf{VidBot} removes this dependency but produces only 3D waypoints. In contrast, our method predicts dense 6-DoF end-effector poses without requiring any human intervention.

\input{tables/real_world_exp}

\subsection{Results and Discussions}
\noindent \textbf{Simulator Benchmark.}
As shown in Tab.~\ref{tab:benchmark}, our method achieves the highest overall success rate of 90.8\%, representing a 14.9 percentage point (pp) improvement over the runner-up~\cite{chen2025vidbot}.
Among the baselines, \textbf{Where2Act} performs well on tasks involving hinge-articulated and portable objects, but exhibits limited generalization to novel objects. 2D affordance prediction methods, \textbf{VRB} and \textbf{RAM}, achieve performance comparable to \textbf{Where2Act} on simpler tasks; however, they struggle significantly on articulated objects requiring complex, curved 3D motions, such as door-opening in \textbf{\texttt{T03}} and \textbf{\texttt{T05}}. These failures stem from oversimplified 2D motion cues, which lead to infeasible linear motions and frequent gripper slips during execution.
3D affordance prediction methods, such as \textbf{GFlow} and \textbf{VidBot}, achieve notable improvements over 2D-based methods, particularly on tasks that require complex motion. Nevertheless, their performance varies substantially between tasks, potentially due to the scale and diversity of their training datasets.

Overall, these comparisons reveal that while 3D affordance representations improve performance on complex manipulation tasks, existing methods still struggle to generalize reliably across objects and viewpoints. In contrast, our approach captures functional similarities between objects through \fieldname~while accounting for geometric and visual differences via an intuitive optimization scheme. As a result, it produces more feasible, scale-aware actions, reducing gripper slip, and achieving a higher success rate.

\begin{figure}[t]
\centering
\includegraphics[width=\linewidth]{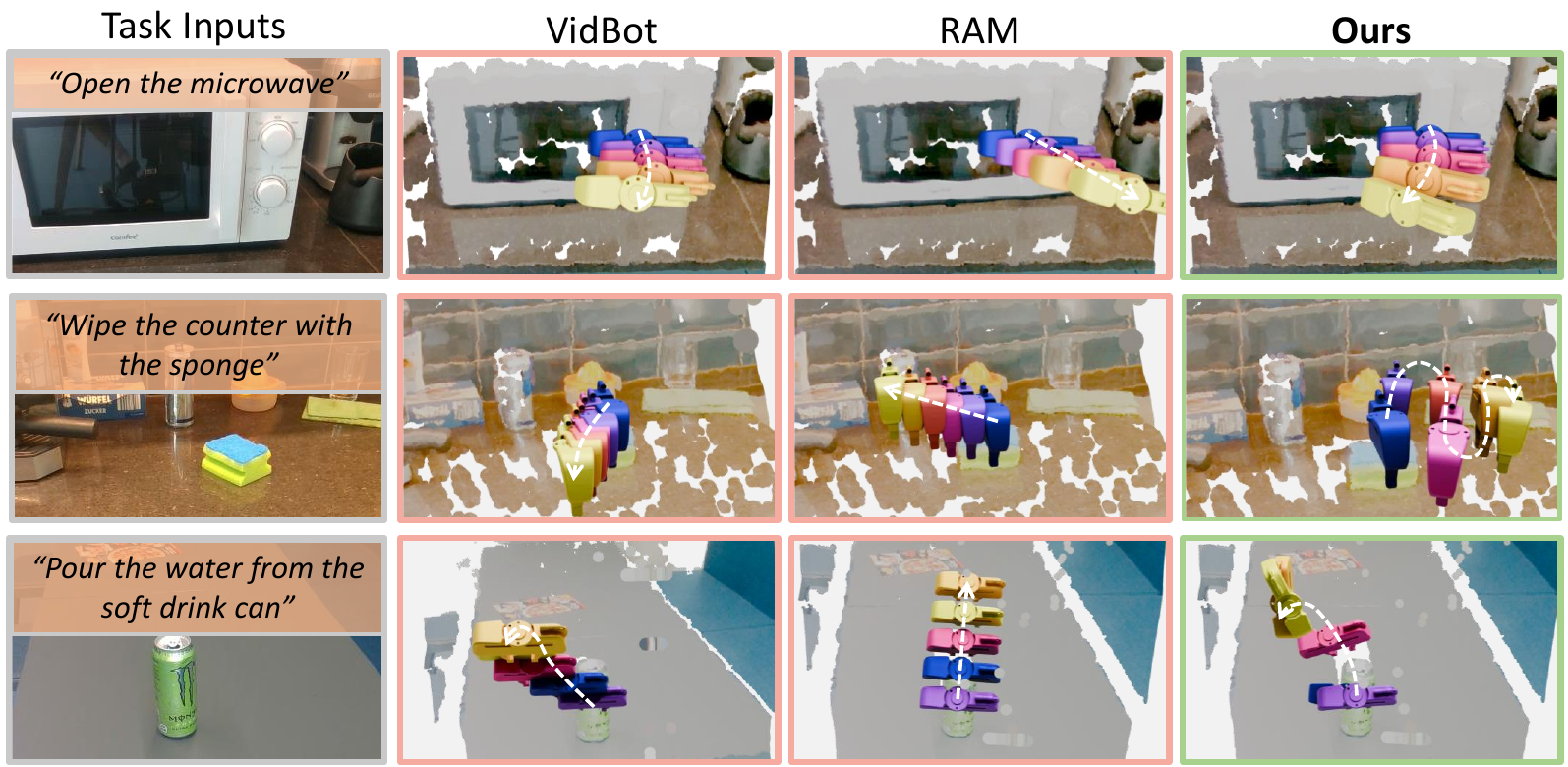}
\vspace{-0.2in}
\caption{Inferred robot gripper actions by different methods.}
\vspace{-0.3in}
\label{fig:realworld_qualitative}
\end{figure}

\noindent \textbf{Real Robot Experiments.}
We evaluated our method on five real-world tasks, performing 10 trials for each and comparing against \textbf{RAM}~(retrieval-based) and \textbf{VidBot}~(prediction-based). As shown in Tab.~\ref{tab:real_robot_qualitative}, our approach achieves an average SR of 76\%. Notably, only our method generated full 6-DoF motions, essential for tasks such as pouring, whereas the baselines were limited to purely translational motions, resulting in potential task failures (Fig.~\ref{fig:realworld_qualitative}).

\vspace{-0.05in}

\subsection{Ablation Studies}
We conducted detailed ablation experiments on a subset of tasks to study the impact of each key component of our pipeline. The results are summarized in Tab.~\ref{tab:ablation}. \textbf{V1} replaces \textsc{NAF} with direct 2D feature-based matching and transfer~\cite{kuang2024ram}. 
\textbf{V2} removes contact-guided initialization and \textbf{V3} disables feature-based ranking. \textbf{V4} bypasses optimization and transfers affordance from the most similar retrieved view, while \textbf{V5} removes the contact-centric refinement stage.

\noindent \textbf{Impact of Neural Representation (V1)}:
We demonstrate the effectiveness of embedding action-related information within a neural function. In \textbf{V1}, performance drops by over 50\%, which is expected as large viewpoint gaps between demonstration and task scenes make direct transfer unreliable. Without a functional representation, the system cannot leverage multimodal optimization for robust affordance alignment, leading to significant degradation.

\noindent \textbf{Impact of Initialization Strategy (V2-V3)}: 
In \textbf{V2}, disabling contact-guided sampling allows infeasible poses to enter the optimization module, increasing susceptibility to local minima and reducing SR by 30\%. 
In \textbf{V3}, removing the ranking module reduces SR by 28.9\%, as relying solely on geometric plausibility often results in suboptimal initialization. This underscores the critical role of feature-informed ranking in guiding the optimizer toward feasible and high-quality poses.

\noindent \textbf{Impact of the Optimization Scheme (V4-V5)}: Disabling the entire optimization (\textbf{V4}) and transferring affordance solely from the most similar views drastically reduces SR to 21.1\%. This sharp drop indicates that sparse 2D visual alignment, without 3D geometric alignment, is insufficient for handling large viewpoint variations due to the gaps between the selected view and the target scene. Removing the final contact-centric refinement (\textbf{V5}) results in a decline in SR to 66.7\%, as the global alignment from the first stage cannot ensure precise alignment in the contact area, where the desired actionable flows reside. This highlights the need for contact-constrained refinement.

\input{tables/ablation_table}

\subsection{Downstream Applications}
We further showcase the versatility of \methodname by applying it to a range of downstream applications.

\noindent \textbf{Long-horizon Manipulation.} 
\methodname can be seamlessly integrated into MLLM-based task and motion planning frameworks~\cite{yang2025LLMTAMP}, enabling flexible and complex manipulation from free-form language instructions. Given an instruction, the MLLM~\cite{openai2024gpt4o} produces a high-level plan as a sequence of sub-tasks. For each step, \methodname retrieves the corresponding \textsc{NAF}, aligns it with the current observation, and transfers the action. As demonstrated in Fig.~\ref{fig:teaser_fig}, the given instruction \textit{"Use the sponge in the drawer to clean the table"} is decomposed into three sub-tasks and executed sequentially, demonstrating our framework's ability to support more complex manipulation tasks.

\begin{figure}[t]
\centering
\includegraphics[width=\linewidth]{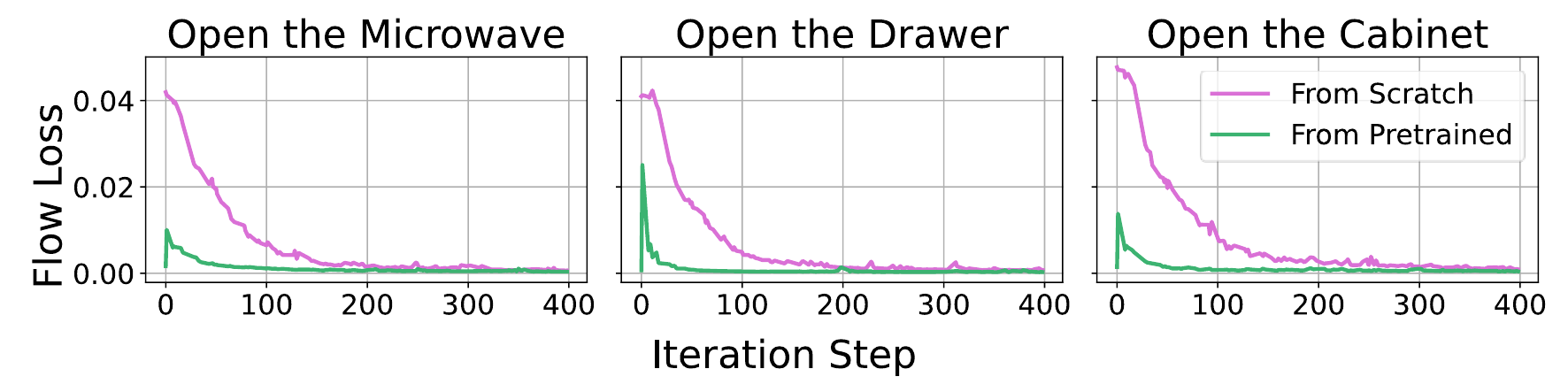}
\vspace{-0.25in}
\caption{Loss curves of the action head when trained from scratch vs. initialized with pretrained weights of the retrieved \textsc{NAF} model.}
\vspace{-0.3in}
\label{fig:flow_loss_comparison}
\end{figure}

\noindent \textbf{Policy Fine-tuning.}
During deployment, the action head ${\color{RubineRed}f_\texttt{act}}$ of the retrieved \textsc{NAF} can be further optimized online for \emph{previously unseen objects}, using observations collected from successful executions. 
We evaluated this procedure in tasks \textbf{\texttt{T01}}, \textbf{\texttt{T03}}, and \textbf{\texttt{T05}} in the same settings as discussed in Sec.~\ref{sec:exp}, achieving absolute SR improvements of 6.7\%, 13.3\%, and 6.7\%, respectively. 
These gains arise because fine-tuning with local motion data helps correct errors from SfM and monocular depth estimation, thereby improving overall performance. 
Moreover, convergence is accelerated around $\times4$ (Fig.~\ref{fig:flow_loss_comparison}): fine-tuning reaches convergence in $\sim$100 steps.
Together, these results further validate that the latent features encoded by $\textsc{NAF}$ provide beneficial representations for action learning.

%% file: tables/quantiative_results.tex
\begin{table*}[t]
\label{exp:sota_table}
\vspace{0.06in}
\centering
\small
\setlength{\tabcolsep}{8.0pt}
 \fontsize{7}{8}\selectfont
\begin{tabular}{@{}r|ccccccccccccc|c@{}}
\toprule
                 &  \textbf{\texttt{T01}}
                 &  \textbf{\texttt{T02}}       
                 &  \textbf{\texttt{T03}}
                 &  \textbf{\texttt{T04}}        
                 &  \textbf{\texttt{T05}}
                 &  \textbf{\texttt{T06}}           
                 &  \textbf{\texttt{T07}}
                 &  \textbf{\texttt{T08}}        
                 &  \textbf{\texttt{T09}}
                 &  \textbf{\texttt{T10}}        
                 &  \textbf{\texttt{T11}}
                 &  \textbf{\texttt{T12}}
                 &  \textbf{\texttt{T13}}
                & \textbf{\texttt{Avg.}}
                 \\
 
\cmidrule(lr){1-1}\cmidrule(lr){2-14} \cmidrule(lr){15-15}
Where2Act~\cite{mo2021where2act}      &     \underline{60.0}   &    40.0      &        33.3     &  \underline{53.3}         &     \textbf{100.0}      &     \textbf{100.0}      &     40.0       &      6.7      &       33.3    &         \underline{40.0}       &    26.7     &    \textbf{100.0}       &    \textbf{100.0}       &   56.4  \\
VRB$^\dagger$~\cite{bahl2023VRB}   &   \underline{60.0}   &    \underline{60.0}     &  40.0    &  40.0         &        13.3         &     \textbf{100.0}     &       0.0     &       0.0   &      53.3      & 10.0  &    \underline{53.3}   &    \underline{93.3}    &    \textbf{100.0}   &    42.6    \\
RAM~\cite{kuang2024ram}  &   46.7    &    \underline{60.0}      &        33.3     &  \underline{53.3}         &    13.3      &     \underline{93.3}     &     33.3       &      $\textbf{93.3}^*$      &       $\text{66.7}^*$     &         $\text{33.3}^*$         &    $\underline{\text{53.3}}^*$      &    80.0 & 80.0 & 58.5      \\
GFlow~\cite{yuan2024generalflow}   &   6.7    &    13.3     & 20.0    &    33.3       &     20.0      &    \textbf{100.0}    &       \underline{93.3}      &       0.0 &      13.3     &   0.0        &    20.0   &    \textbf{100.0}    &    \underline{93.3}    &   39.6  \\
VidBot~\cite{chen2025vidbot}      &   46.7   &    \textbf{100.0}     & \underline{73.3}    &  \textbf{100.0}    &     26.7      &     \textbf{100.0}     &     33.3      &     \underline{86.7}   &      \underline{73.3}      &   \textbf{93.3}        &    46.7   &    \textbf{100.0}    &    \textbf{100.0}    &    \underline{75.9} \\

\textbf{Ours} & \textbf{73.3} & \textbf{100.0} &  \textbf{80.0} & \textbf{100.0}
& \underline{80.0} & \underline{93.3} & \textbf{100.0} & \textbf{93.3}  & 
\textbf{86.7}      &   \textbf{93.3}         &    \textbf{73.3}  & \textbf{100.0} & \textbf{100.0} & \textbf{90.8}\\

\bottomrule
\end{tabular}
\caption{Quantitative results on tasks evaluated in simulators on success rate (\%). \textbf{\texttt{T01}}: Open drawer \textbf{\texttt{T02}}: Close drawer, \textbf{\texttt{T03}}: Open microwave,  \textbf{\texttt{T04}}: Close microwave, \textbf{\texttt{T05}}: Open hinge cabinet,  \textbf{\texttt{T06}}: Close hinge cabinet, \textbf{\texttt{T07}}: Open dishwasher,  \textbf{\texttt{T08}}: Open slide cabinet, \textbf{\texttt{T09}}: Close slide cabinet,   \textbf{\texttt{T10}}: Close laptop lid, \textbf{\texttt{T11}} Put down toilet seat: \textbf{\texttt{T12}}: Pick up cup, \textbf{\texttt{T13}}: Pick up bottle. $^\text{*}$ uses self-collected data for novel tasks. $^\dagger$ uses strategy from \cite{kuang2024ram} to lift affordance to 3D. }

\vspace{-0.3in}

\label{tab:benchmark}.  
\end{table*}

%% file: tables/real_world_exp.tex
\begin{table}[t]
\centering
\small
\setlength{\tabcolsep}{9.5pt}
\fontsize{7}{8}\selectfont
 
\begin{tabular}{@{}r|ccccc|c@{}}
\toprule
                 &  \textbf{\texttt{RT01}}
                 &  \textbf{\texttt{RT02}}       
                 &  \textbf{\texttt{RT03}}
                 &  \textbf{\texttt{RT04}}        
                 &  \textbf{\texttt{RT05}}
                 &  \textbf{\texttt{Avg.}}
                 \\
 
\cmidrule(lr){1-1} \cmidrule(lr){2-7}
RAM             & 40.0 & 20.0 & 60.0         & $\underline{\text{0.0}}^*$        & $\text{40.0}^*$      &  32.0       \\
VidBot    &  \underline{60.0}         &  \underline{40.0 }        & \textbf{100.0} & \underline{0.0}   & \underline{50.0} & \underline{50.0}    \\
\textbf{Ours}  &  \textbf{70.0}   & \textbf{70.0}     & \underline{90.0}   & \textbf{80.0}      &  \textbf{70.0}    & \textbf{76.0}       \\
\bottomrule
\end{tabular}
\caption{Real-world experiment results on selected tasks. \textbf{\texttt{RT01}}: Open drawer, \textbf{\texttt{RT02}}: Open microwave, \textbf{\texttt{RT03}}: Close microwave, \textbf{\texttt{RT04}}: Pour water, \textbf{\texttt{RT05}}: Wiping table with sponge. $^\text{*}$: Use self-collected data for retrieval. }
\label{tab:real_robot_qualitative}
\vspace{-0.3in}

\end{table}

%% file: tables/ablation_table.tex
\begin{table}[t]
\centering
\small
\setlength{\tabcolsep}{2.5pt}
\fontsize{7}{8}\selectfont
\vspace{0.06in}
\begin{tabular}{@{}r|cccccc|c@{}}
\toprule
                 &  \textbf{\texttt{AT01}}
                 &  \textbf{\texttt{AT02}}       
                 &  \textbf{\texttt{AT03}}
                 &  \textbf{\texttt{AT04}}        
                 &  \textbf{\texttt{AT05}}
                 &  \textbf{\texttt{AT06}}
                 &  \textbf{\texttt{Avg.}}
                 \\
 
\cmidrule(lr){1-1} \cmidrule(lr){2-7} \cmidrule(lr){8-8}

\textbf{Ours [Full Model]}               &  \textbf{73.3}   & \textbf{100.0}     & \textbf{80.0}   & \textbf{100.0}      &  \underline{93.3}       & \textbf{73.3}   & \textbf{86.7}       \\
\cmidrule(lr){1-8}  
w/o Neural Function \textbf{[V1]}              &\underline{66.7} & 26.7 & \underline{73.3}               & 0.0      &   6.7      & 6.7             &  30.0       \\
w/o Contact Sampling \textbf{[V2]}   &  53.3          &  13.3             & 26.7 & \underline{73.3}  &   \textbf{100.0}   &\textbf{ 73.3} & 56.6     \\
w/o Viewpoint Ranking \textbf{[V3]}           & \underline{66.7} & 33.3 & 0.0               & \textbf{100.0}      &   \textbf{100.0}      & 46.7             &  57.8      \\
w/o Afford. Optimization \textbf{[V4]}      &\textbf{73.3}& 0.0 & 0.0               & 26.7    &   20.0      & 6.7             &  21.1       \\
w/o Contact Refinement \textbf{[V5]}   &  \underline{66.7}    &  \underline{66.7}           & 33.3               & 66.7    &   \textbf{100.0}      & \underline{66.7} &  \underline{66.7}   \\

\bottomrule
\end{tabular}
\caption{Ablation results on 6 selected tasks. \textbf{\texttt{AT01}}: Open drawer, \textbf{\texttt{AT02}}: Close drawer, \textbf{\texttt{AT03}}: Open microwave,  \textbf{\texttt{AT04}}: Close microwave, \textbf{\texttt{AT05}}: Open slide cabinet. \textbf{\texttt{AT06}}: Put down toilet seat. $^\dagger$: Use strategy from \cite{kuang2024ram} to lift affordance to 3D.}
\label{tab:ablation}
\vspace{-0.3in}

\end{table}

%% file: sections/06_conclusion.tex
\section{Conclusion} \label{sec:conc}
We introduce \methodname, a framework for zero-shot robotic manipulation that learns continuous object-centric action functions from monocular videos and transfers them to novel objects and viewpoints. 
Central to our approach is the \emph{Neural Affordance Function} that jointly encodes geometry, features, contact priors, and point flows.
With an optimization-based transfer pipeline, \methodname enables generalization without relying on large-scale teleoperated demonstrations. Our pipeline resolves the inherent ambiguity of 2D cues in prior works, improves data efficiency, and bridges viewpoint discrepancies. 
With a few video demonstrations, our framework consistently surpasses strong baselines on diverse household tasks, both in simulation and in the real world, demonstrating its strong scalability and generalizability.

\noindent \textbf{Limitations and Future Work.} 
Our method leverages recent 3D AIGC techniques~\cite{wonder3d_long2023} to infer knowledge from unseen views; however, artifacts in these generated views can negatively affect downstream action prediction. Moreover, inference relies on iterative optimization and, therefore, cannot yet satisfy strict real-time latency requirements. Future work will investigate acceleration strategies to enhance deployment efficiency, including vectorized optimization~\cite{kong2023vmap} and second-order solvers~\cite{höllein20253dgslm}.

%% file: sections/07_ack.tex
\section*{Acknowledgement} \label{sec:acknowledgement}
This work was supported by the Technical University of Munich (TUM) and the State of Bavaria through the REACT project, the TUM Georg Nemetschek Institute via the SPAICR project, Munich Center for Machine Learning (MCML) and ETH Zurich.

%% file: reference.bib
@string{icra = {IEEE Int. Conf. on Robotics and Automation (ICRA)}}

@string{cvpr = {CVPR}}

@string{iccv = {ICCV}}

@string{eccv = {ECCV}}

@string{tog = {ACM TOG}}

@article{amir2021deep,
 author = {Shir Amir and Yossi Gandelsman and Shai Bagon and Tali Dekel},
 journal = {ECCVW What is Motion For?},
 title = {Deep ViT Features as Dense Visual Descriptors},
 year = {2022}
}

@article{patel2025robotic,
  title={Robotic Manipulation by Imitating Generated Videos Without Physical Demonstrations},
  author={Patel, Shivansh and Mohan, Shraddhaa and Mai, Hanlin and Jain, Unnat and Lazebnik, Svetlana and Li, Yunzhu},
  journal={arXiv preprint arXiv:2507.00990},
  year={2025}
}

@inproceedings{Austin2015AffordanceDet,
 author = {Myers, Austin and Teo, Ching L. and Fermüller, Cornelia and Aloimonos, Yiannis},
 booktitle = {ICRA},
 doi = {10.1109/ICRA.2015.7139369},
 number = {},
 pages = {1374-1381},
 title = {Affordance detection of tool parts from geometric features},
 volume = {},
 year = {2015}
}

@article{bahl2022human,
 author = {Bahl, Shikhar and Gupta, Abhinav and Pathak, Deepak},
 journal = {arXiv preprint arXiv:2207.09450},
 title = {Human-to-robot imitation in the wild},
 year = {2022}
}

@inproceedings{bahl2023affordances,
 author = {Bahl, Shikhar and Mendonca, Russell and Chen, Lili and Jain, Unnat and Pathak, Deepak},
 booktitle = {CVPR},
 pages = {13778--13790},
 title = {Affordances from human videos as a versatile representation for robotics},
 year = {2023}
}

@inproceedings{bahl2023VRB,
 author = {Bahl, Shikhar and Mendonca, Russell and Chen, Lili and Jain, Unnat and Pathak, Deepak},
 booktitle = {CVPR},
 title = {Affordances from Human Videos as a Versatile Representation for Robotics},
 year = {2023}
}

@article{BeslSVD,
 author = {Besl, P.J. and McKay, Neil D.},
 doi = {10.1109/34.121791},
 journal = {TAPMI},
 number = {2},
 pages = {239-256},
 title = {A method for registration of 3-D shapes},
 volume = {14},
 year = {1992}
}

@misc{bharadhwaj2024gen2act,
 archiveprefix = {arXiv},
 author = {Homanga Bharadhwaj and Debidatta Dwibedi and Abhinav Gupta and Shubham Tulsiani and Carl Doersch and Ted Xiao and Dhruv Shah and Fei Xia and Dorsa Sadigh and Sean Kirmani},
 note = {arXiv:2409.16283},
 primaryclass = {cs.RO},
 title = {Gen2Act: Human Video Generation in Novel Scenarios enables Generalizable Robot Manipulation},
 year = {2024}
}

@article{bharadhwaj2024track2act,
 author = {Bharadhwaj, Homanga and Mottaghi, Roozbeh and Gupta, Abhinav and Tulsiani, Shubham},
 journal = {arXiv preprint arXiv:2405.01527},
 title = {Track2Act: Predicting Point Tracks from Internet Videos enables Diverse Zero-shot Robot Manipulation},
 year = {2024}
}

@inproceedings{chen2024funcgrasp,
 author = {Chen, Hanzhi and Xu, Binbin and Leutenegger, Stefan},
 booktitle = {ICRA},
 organization = {IEEE},
 pages = {1900--1906},
 title = {FuncGrasp: Learning Object-Centric Neural Grasp Functions from Single Annotated Example Object},
 year = {2024}
}

@article{chen2025vidbot,
 author = {Chen, Hanzhi and Sun, Boyang and Zhang, Anran and Pollefeys, Marc and Leutenegger, Stefan},
 journal = {CVPR},
 title = {VidBot: Learning Generalizable 3D Actions from In-the-Wild 2D Human Videos for Zero-Shot Robotic Manipulation},
 year = {2025}
}

@inproceedings{chuang2018learning,
 author = {Chuang, Ching-Yao and Li, Jiaman and Torralba, Antonio and Fidler, Sanja},
 booktitle = {CVPR},
 pages = {975--983},
 title = {Learning to act properly: Predicting and explaining affordances from images},
 year = {2018}
}

@inproceedings{do2018affordancenet,
 author = {Do, Thanh-Toan and Nguyen, Anh and Reid, Ian},
 booktitle = {ICRA},
 organization = {IEEE},
 pages = {5882--5889},
 title = {Affordancenet: An end-to-end deep learning approach for object affordance detection},
 year = {2018}
}

@inproceedings{Drawer2025xia,
 author = {Hongchi Xia and Entong Su and Marius Memmel and Arhan Jain and Raymond Yu and Numfor Mbiziwo-Tiapo and Ali Farhadi and Abhishek Gupta and Shenlong Wang and Wei-Chiu Ma},
 title = {DRAWER: Digital Reconstruction and Articulation With Environment Realism},
 year = {2025},
booktitle={CVPR}
}

@article{fang2023anygrasp,
 author = {Fang, Hao-Shu and Wang, Chenxi and Fang, Hongjie and Gou, Minghao and Liu, Jirong and Yan, Hengxu and Liu, Wenhai and Xie, Yichen and Lu, Cewu},
 journal = {T-RO},
 title = {AnyGrasp: Robust and Efficient Grasp Perception in Spatial and Temporal Domains},
 year = {2023}
}

@article{florence2018dense,
 author = {Florence, Peter R and Manuelli, Lucas and Tedrake, Russ},
 journal = {CoRL},
 title = {Dense object nets: Learning dense visual object descriptors by and for robotic manipulation},
 year = {2018}
}

@misc{höllein20253dgslm,
 archiveprefix = {arXiv},
 author = {Lukas Höllein and Aljaž Božič and Michael Zollhöfer and Matthias Nießner},
 note = {arXiv:2409.12892},
 primaryclass = {cs.CV},
 title = {3DGS-LM: Faster Gaussian-Splatting Optimization with Levenberg-Marquardt},
 year = {2025}
}

@article{james2019rlbench,
 author = {James, Stephen and Ma, Zicong and Rovick Arrojo, David and Davison, Andrew J.},
 journal = {RA-L},
 title = {RLBench: The Robot Learning Benchmark \& Learning Environment},
 year = {2020}
}

@article{jones2007imitation,
 author = {Jones, Susan S},
 journal = {Psychological science},
 number = {7},
 pages = {593--599},
 publisher = {SAGE Publications Sage CA: Los Angeles, CA},
 title = {Imitation in infancy: The development of mimicry},
 volume = {18},
 year = {2007}
}

@misc{ju2024roboabc,
 archiveprefix = {arXiv},
 author = {Yuanchen Ju and Kaizhe Hu and Guowei Zhang and Gu Zhang and Mingrun Jiang and Huazhe Xu},
 note = {arXiv:2401.07487},
 primaryclass = {cs.RO},
 title = {Robo-ABC: Affordance Generalization Beyond Categories via Semantic Correspondence for Robot Manipulation},
 year = {2024}
}

@article{kerbl3Dgaussians,
 author = {Kerbl, Bernhard and Kopanas, Georgios and Leimk{\"u}hler, Thomas and Drettakis, George},
 journal = {ACM Transactions on Graphics},
 month = {July},
 number = {4},
 title = {3D Gaussian Splatting for Real-Time Radiance Field Rendering},
 volume = {42},
 year = {2023}
}

@article{kingma2014adam,
 author = {Kingma, Diederik P and Ba, Jimmy},
 journal = {arXiv preprint arXiv:1412.6980},
 title = {Adam: A method for stochastic optimization},
 year = {2014}
}

@inproceedings{kirillov2023segment,
 author = {Kirillov, Alexander and Mintun, Eric and Ravi, Nikhila and Mao, Hanzi and Rolland, Chloe and Gustafson, Laura and Xiao, Tete and Whitehead, Spencer and Berg, Alexander C and Lo, Wan-Yen and others},
 booktitle = {ICCV},
 pages = {4015--4026},
 title = {Segment anything},
 year = {2023}
}

@misc{kong2023vmap,
 archiveprefix = {arXiv},
 author = {Xin Kong and Shikun Liu and Marwan Taher and Andrew J. Davison},
 note = {arXiv:2302.01838},
 primaryclass = {cs.CV},
 title = {vMAP: Vectorised Object Mapping for Neural Field SLAM},
 year = {2023}
}

@misc{kuang2024ram,
 author = {Kuang, Yuxuan and Ye, Junjie and Geng, Haoran and Mao, Jiageng and Deng, Congyue and Guibas, Leonidas and Wang, He and Wang, Yue},
 journal = {arXiv preprint},
 title = {RAM: Retrieval-Based Affordance Transfer for Generalizable Zero-Shot Robotic Manipulation},
 year = {2024},
note={arXiv:2407.04689}
}

@inproceedings{li2024_MegaSaM,
      title={MegaSaM: Accurate, Fast, and Robust Structure and Motion from Casual Dynamic Videos}, 
      author={Zhengqi Li and Richard Tucker and Forrester Cole and Qianqian Wang and Linyi Jin and Vickie Ye and Angjoo Kanazawa and Aleksander Holynski and Noah Snavely},
      booktitle = {CVPR},
      year={2024},
}

@misc{liang2024dreamitate,
 archiveprefix = {arXiv},
 author = {Junbang Liang and Ruoshi Liu and Ege Ozguroglu and Sruthi Sudhakar and Achal Dave and Pavel Tokmakov and Shuran Song and Carl Vondrick},
 note = {arXiv:2406.16862},
 primaryclass = {cs.RO},
 title = {Dreamitate: Real-World Visuomotor Policy Learning via Video Generation},
 year = {2024}
}

@inproceedings{mo2021where2act,
 author = {Mo, Kaichun and Guibas, Leonidas J and Mukadam, Mustafa and Gupta, Abhinav and Tulsiani, Shubham},
 booktitle = {ICCV},
 pages = {6813--6823},
 title = {Where2act: From pixels to actions for articulated 3d objects},
 year = {2021}
}

@article{mueller2022instant,
 author = {M{\"u}ller, Thomas and Evans, Alex and Schied, Christoph and Keller, Alexander},
 journal = {ACM ToG},
 publisher = {ACM New York, NY, USA},
 title = {{Instant Neural Graphics Primitives with a Multiresolution Hash Encoding}},
 year = {2022}
}

@misc{nair2022r3m,
 author = {Nair, Suraj and Rajeswaran, Aravind and Kumar, Vikash and Finn, Chelsea and Gupta, Abhinav},
 title = {R3m: A universal visual representation for robot manipulation},
 year = {2022},
 note={arXiv:2203.12601}
}

@misc{openai2024gpt4o,
 archiveprefix = {arXiv},
 author = {OpenAI},
 note = {arXiv:2410.21276},
 primaryclass = {cs.CL},
 title = {GPT-4o System Card},
 year = {2024}
}

@misc{oquab2023dinov2,
 author = {Oquab, Maxime and Darcet, Timoth{\'e}e and Moutakanni, Th{\'e}o and Vo, Huy and Szafraniec, Marc and Khalidov, Vasil and Fernandez, Pierre and Haziza, Daniel and Massa, Francisco and El-Nouby, Alaaeldin and others},
 title = {Dinov2: Learning robust visual features without supervision},
 year = {2023},
 note={arXiv:2304.07193}
}

@article{papagiannis2024r+,
 author = {Georgios Papagiannis and Norman Di Palo and Pietro Vitiello and Edward Johns},
 journal = {ICRA},
 title = {R+X: Retrieval and Execution from Everyday Human Videos},
 year = {2025}
}

@misc{piccinelli2025unidepthv2,
 archiveprefix = {arXiv},
 author = {Luigi Piccinelli and Christos Sakaridis and Yung-Hsu Yang and Mattia Segu and Siyuan Li and Wim Abbeloos and Luc Van Gool},
 note = {arXiv:2502.20110},
 primaryclass = {cs.CV},
 title = {{U}ni{D}epth{V2}: Universal Monocular Metric Depth Estimation Made Simpler},
 year = {2025}
}

@inproceedings{qin2022dexmv,
 author = {Qin, Yuzhe and Wu, Yueh-Hua and Liu, Shaowei and Jiang, Hanwen and Yang, Ruihan and Fu, Yang and Wang, Xiaolong},
 booktitle = {ECCV},
 organization = {Springer},
 pages = {570--587},
 title = {Dexmv: Imitation learning for dexterous manipulation from human videos},
 year = {2022}
}

@inproceedings{rashid2023lerftogo,
 author = {Adam Rashid and Satvik Sharma and Chung Min Kim and Justin Kerr and Lawrence Yunliang Chen and Angjoo Kanazawa and Ken Goldberg},
 booktitle = {CoRL},
 title = {Language Embedded Radiance Fields for Zero-Shot Task-Oriented Grasping},
 year = {2023}
}

@inproceedings{shaw2023videodex,
 author = {Shaw, Kenneth and Bahl, Shikhar and Pathak, Deepak},
 booktitle = {CoRL},
 organization = {PMLR},
 pages = {654--665},
 title = {Videodex: Learning dexterity from internet videos},
 year = {2023}
}

@inproceedings{shen2023F3RM,
 author = {Shen, William and Yang, Ge and Yu, Alan and Wong, Jansen and Kaelbling, Leslie Pack and Isola, Phillip},
 booktitle = {CoRL},
 title = {Distilled Feature Fields Enable Few-Shot Language-Guided Manipulation},
 year = {2023}
}

@article{smith2019avid,
 author = {Smith, Laura and Dhawan, Nikita and Zhang, Marvin and Abbeel, Pieter and Levine, Sergey},
 journal = {arXiv preprint arXiv:1912.04443},
 title = {Avid: Learning multi-stage tasks via pixel-level translation of human videos},
 year = {2019}
}

@misc{tapip3d,
 author = {Zhang, Bowei and Ke, Lei and Harley, Adam W and Fragkiadaki, Katerina},
 title = {TAPIP3D: Tracking Any Point in Persistent 3D Geometry},
 year = {2025},
 note={arXiv:2504.14717}
}

@article{wang2021neus,
 author = {Peng Wang and Lingjie Liu and Yuan Liu and Christian Theobalt and Taku Komura and Wenping Wang},
 journal = {NeurIPS},
 title = {NeuS: Learning Neural Implicit Surfaces by Volume Rendering for Multi-view Reconstruction},
 year = {2021}
}

@article{wang2022adaafford,
 author = {Wang, Yian and Wu, Ruihai and Mo, Kaichun and Ke, Jiaqi and Fan, Qingnan and Guibas, Leonidas and Dong, Hao},
 journal = {ECCV},
 title = {{AdaAfford}: Learning to Adapt Manipulation Affordance for 3D Articulated Objects via Few-shot Interactions},
 year = {2022}
}

@misc{wang2024d3fields,
 archiveprefix = {arXiv},
 author = {Yixuan Wang and Mingtong Zhang and Zhuoran Li and Tarik Kelestemur and Katherine Driggs-Campbell and Jiajun Wu and Li Fei-Fei and Yunzhu Li},
 note = {arXiv:2309.16118},
 primaryclass = {cs.RO},
 title = {D$^3$Fields: Dynamic 3D Descriptor Fields for Zero-Shot Generalizable Rearrangement},
 year = {2024}
}

@article{wang2024dexcap,
 author = {Wang, Chen and Shi, Haochen and Wang, Weizhuo and Zhang, Ruohan and Fei-Fei, Li and Liu, C Karen},
 journal = {arXiv preprint arXiv:2403.07788},
 title = {Dexcap: Scalable and portable mocap data collection system for dexterous manipulation},
 year = {2024}
}

@misc{wang2024sparsedff,
 archiveprefix = {arXiv},
 author = {Qianxu Wang and Haotong Zhang and Congyue Deng and Yang You and Hao Dong and Yixin Zhu and Leonidas Guibas},
 note = {arXiv:2310.16838},
 primaryclass = {cs.RO},
 title = {SparseDFF: Sparse-View Feature Distillation for One-Shot Dexterous Manipulation},
 year = {2024}
}

@misc{wonder3d_long2023,
 author = {Long, Xiaoxiao and Guo, Yuan-Chen and Lin, Cheng and Liu, Yuan and Dou, Zhiyang and Liu, Lingjie and Ma, Yuexin and Zhang, Song-Hai and Habermann, Marc and Theobalt, Christian and others},
 journal = {arXiv preprint arXiv:2310.15008},
 note = {arXiv:2310.15008},
 title = {Wonder3D: Single Image to 3D using Cross-Domain Diffusion},
 year = {2023}
}

@misc{wu2022vatmart,
 archiveprefix = {arXiv},
 author = {Ruihai Wu and Yan Zhao and Kaichun Mo and Zizheng Guo and Yian Wang and Tianhao Wu and Qingnan Fan and Xuelin Chen and Leonidas Guibas and Hao Dong},
 note = {arXiv:2106.14440},
 primaryclass = {cs.CV},
 title = {VAT-Mart: Learning Visual Action Trajectory Proposals for Manipulating 3D ARTiculated Objects},
 year = {2022}
}

@article{xiao2022masked,
 author = {Xiao, Tete and Radosavovic, Ilija and Darrell, Trevor and Malik, Jitendra},
 journal = {arXiv preprint arXiv:2203.06173},
 title = {Masked visual pre-training for motor control},
 year = {2022}
}

@inproceedings{xu2024flow,
 author = {Mengda Xu and Zhenjia Xu and Yinghao Xu and Cheng Chi and Gordon Wetzstein and Manuela Veloso and Shuran Song},
 booktitle = {CoRL},
 title = {Flow as the Cross-domain Manipulation Interface},
 year = {2024}
}

@misc{xu2024instantmesh,
 author = {Xu, Jiale and Cheng, Weihao and Gao, Yiming and Wang, Xintao and Gao, Shenghua and Shan, Ying},
 journal = {arXiv preprint arXiv:2404.07191},
 note = {arXiv:2404.07191},
 title = {InstantMesh: Efficient 3D Mesh Generation from a Single Image with Sparse-view Large Reconstruction Models},
 year = {2024}
}

@article{yuan2024generalflow,
 author = {Yuan, Chengbo and Wen, Chuan and Zhang, Tong and Gao, Yang},
 journal = {CoRL},
 organization = {PMLR},
 title = {General flow as foundation affordance for scalable robot learning},
 year = {2024}
}

@misc{ze2024gnfactor,
 archiveprefix = {arXiv},
 author = {Yanjie Ze and Ge Yan and Yueh-Hua Wu and Annabella Macaluso and Yuying Ge and Jianglong Ye and Nicklas Hansen and Li Erran Li and Xiaolong Wang},
 note = {arXiv:2308.16891},
 primaryclass = {cs.RO},
 title = {GNFactor: Multi-Task Real Robot Learning with Generalizable Neural Feature Fields},
 year = {2024}
}

@INPROCEEDINGS{yang2025LLMTAMP,
  author={Yang, Zhutian and Garrett, Caelan and Fox, Dieter and Lozano-Pérez, Tomás and Kaelbling, Leslie Pack},
  booktitle={ICRA}, 
  title={Guiding Long-Horizon Task and Motion Planning with Vision Language Models}, 
  year={2025},
  volume={},
  number={},
  pages={16847-16853}
}
